%% file: root.tex
\def\eg{\emph{e.g.}{}} 
\def\ie{\emph{i.e.}}

\def\etal{\emph{et al.}}

\documentclass[10pt,conference,a4paper]{IEEEtran}
\usepackage[pagebackref=true,breaklinks=true,colorlinks,bookmarks=false]{hyperref}
\usepackage{cite}

\ifCLASSINFOpdf
  \usepackage[pdftex]{graphicx}
\else
\fi
\usepackage{hyperref}
\usepackage{booktabs}
\usepackage{multirow}
\usepackage{amsmath}
\usepackage{amsfonts}
\usepackage{url}

\begin{document}
%
\title{Cascade Attention Guided Residue Learning GAN for Cross-Modal Translation}

\author{\IEEEauthorblockN{Bin Duan}
\IEEEauthorblockA{Illinois Institute of Technology, USA\\
bduan2@hawk.iit.edu}
\\
\IEEEauthorblockN{Hugo Latapie}
\IEEEauthorblockA{Cisco, USA\\
hlatapie@cisco.com}
\and
\IEEEauthorblockN{Wei Wang}
\IEEEauthorblockA{
University of Trento, Italy\\
wei.wang@unitn.it}
\and
\IEEEauthorblockN{Hao Tang}
\IEEEauthorblockA{University of Trento, Italy\\
hao.tang@unitn.it}
\\
\IEEEauthorblockN{Yan Yan}
\IEEEauthorblockA{Illinois Institute of Technology, USA\\
yyan34@iit.edu}}

\maketitle

\input{0Abstract}


\ifCLASSOPTIONpeerreview
\begin{center} \bfseries EDICS Category: 3-BBND \end{center}
\fi
%
\IEEEpeerreviewmaketitle

\input{1Introduction.tex}
\input{2Related.tex}
\input{3Approach.tex}
\input{4Experiments.tex}
\input{5Conclusion.tex}

\small{
\bibliographystyle{IEEEtran}
\bibliography{egbib}
}

\end{document}

%% file: 0Abstract.tex
\begin{abstract}
Since we were babies, we intuitively develop the ability to correlate the input from different cognitive sensors such as vision, audio, and text. However, in machine learning, this cross-modal learning is a nontrivial task because different modalities have no homogeneous properties. Previous works discover that there should be bridges among different modalities. From a neurology and psychology perspective, humans have the capacity to link one modality with another one, \eg{}, associating a picture of a bird with the only hearing of its singing and vice versa. Is it possible for machine learning algorithms to recover the scene given the audio signal?

In this paper, we propose a novel \textbf{C}ascade \textbf{A}ttention-Guided \textbf{R}esidue GAN (CAR-GAN), aiming at reconstructing the scenes given the corresponding audio signals.
Particularly, we present a residue module to mitigate the gap between different modalities progressively. Moreover, a cascade attention guided network with a novel classification loss function is designed to tackle the cross-modal learning task. Our model keeps consistency in the high-level semantic label domain and is able to balance two different modalities.  The experimental results demonstrate that our model achieves the state-of-the-art cross-modal audio-visual generation on the challenging Sub-URMP dataset.
\end{abstract}

%% file: 1Introduction.tex
\section{Introduction}
\label{sec:intro}
\par Cross-modal learning involves multiple modalities, aims at learning knowledge from one modality to facilitate the tasks (\eg{}, retrieval and generation) from another correlated modality. 
Cross-modal learning gains long-lasting interest in multimedia. 
Recently, with the increasing popularity of Generative Adversarial Networks (GANs) \cite{goodfellow2014generative}, cross-modal research is not only limited to retrieval~\cite{pereira2014role,rasiwasia2010new}~but also makes the cross-modal generation possible, such as text-to-image \cite{park2018mc,verma2014im2text}, image-to-image \cite{isola2017image,zhu2017unpaired,tang2020unified,liu2020exocentric}, story visualization and generation \cite{li2018storygan,qiao2019mirrorgan}. 
Recently, radio signals~\cite{zhao2018through} have also been successfully applied to human pose prediction. The radio signals which are a form of waves that are robust to occlusions so that it can predict human poses behind the wall. Another special wave is an audio signal, which has also been explored to reconstruct the scene \cite{chen2017deep,hao2018cmcgan,rouditchenko2019self,ngiam2011multimodal,oh2019speech2face}.

\par Generating images from audios using GANs was first described by Chen \etal{} \cite{chen2017deep} where they introduce a conditional GANs (cGANs)~\cite{mirza2014conditional} model to tackle the problem. 
Later, Hao \etal{} presented the Cross-Modal Cycle GAN (CMCGAN)~\cite{hao2018cmcgan} to solve the cross-modal visual-audio mutual generation problem. 
Although this paper conducts an interesting exploration, we still observe unsatisfactory artifacts and missing contents in the generated images, which are due to several reasons.
First, even if previous works \cite{chen2017deep,hao2018cmcgan} showed there were truly some connections between audio and visual modalities, there is still a huge gap between different modalities, \eg{}, the sound of the wind blowing trees and the image of shaking leaves. Thus, without prior knowledge about this scenario, it is hard to associate them together. 
Second, a random latent vector was employed to assist the learning process. They tried to represent the properties of the input audios accompanied by some manually defined random latent vectors.
However, we argue that these latent vectors cannot represent the information of the audios accurately since they are random Gaussian noises, \ie{}, they are not directly withdrawn from the audios. Consequently, we avoid employing random latent vectors in our designed model.

\begin{figure*}[t]
	\centering
	\includegraphics[width=0.9\linewidth,
	height=5cm]{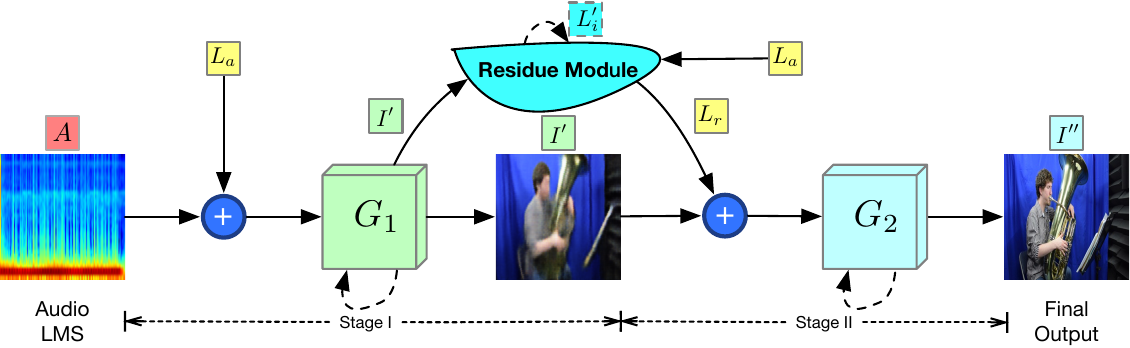}
	\caption{Overview of CAR-GAN framework. Generator $G_1$ takes audio LMS ($A$) and its class label ($L_a$) as inputs to generate coarse image $I'$. Generator $G_2$ takes $I'$ and the output of our specific-designed residue module ($L_r$) as inputs to synthesize fine-grained image $I''$. Note that to detect the most distinguished part in different modalities, we introduce the self-attention mechanism to both generators. Two generators of different stages are jointly optimized in an end-to-end fashion that aims at enjoying the mutually improved benefits from different modalities, \ie{}, audio and image.~\textcircled{+}~denotes channel-wise concatenation.}
	\label{fig:framework}
\end{figure*}

\par Based on the above observations, in this paper, we propose a novel \textbf{C}ascade \textbf{A}ttention-Guided \textbf{R}esidue GAN (CAR-GAN) to handle the audio-to-image translation task. 
The proposed CAR-GAN contains two-generation stages and the overall framework of CAR-GAN is depicted in Fig.~\ref{fig:framework}, in the first stage, the Log-amplitude of Mel-Spectrum (LMS) \cite{deng2004enhancement} image $A$ (a representation of the raw audio) concatenated with ${L_a}$ (one-hot class label for the audio) is fed into the first self-attention guided generator ($G_1$) and $G_1$ outputs a coarse result $I'$.
Different from previous works \cite{chen2017deep,hao2018cmcgan}, we deprecate the latent vector in our model. 
Note that to efficiently model relationships between widely separated spatial regions, we introduce self-attention to both generator and discriminator.

The coarse output $I'$ from the first generation network is taken as input to the proposed residue module to obtain the corresponding class label vector $L_i'$, which is a high-level semantic class label for the coarse output $I'$ produced by the embedded classifier in our proposed residue module. The label $L_i$ reflects how realistic the first generator is by measuring the embedding in the high-level class label space.
Then we subtract $L_i'$ from label $L_a$ to obtain the residue class label vector $L_r$. $L_r$ reflects the discrepancy between the generated image $I'$ and the audio in the semantic label domain based on the coarse output $I'$, which is corresponding to the discrepancy between $L_i'$ and $L_a$. By taking means of $L_r$, our proposed model makes compensation for the information lost in the first stage and generates more realistic results in the second stage.
Next, the coarse output $I'$ from the first stage, together with the residue class label $L_r$, are input into the second stage network and generate more fine-grained final results.
The intuition behind the residue module is that the second generator $G_2$ can flexibly preserve the similarities between audio and image space and only model the differences when it is necessary, which can be regarded as a progressive generation strategy.
Finally, to optimize the proposed CAR-GAN in an end-to-end fashion, the cascaded residue classification loss is further used to generate more realistic images and preserve the consistency between two stages.
It is worth noting that, the classifier in our proposed residue module is pre-trained with the real images from the Sub-URMP dataset. During the training of our CAR-GAN model, the classifier parameters are fixed, and only the gradients of the synthesized images will be backpropagated to guide the generator to synthesize images with correct semantic labels. In this way, the label consistency between the synthesized images and the audio labels could be preserved.
Through extensive experimental evaluations, we demonstrate that CAR-GAN produces better results than the baselines such as S2IC \cite{chen2017deep} and CMCGAN \cite{hao2018cmcgan}.

\par Overall, the contributions of this paper are summarized as follows:
\begin{itemize}
	\item A novel Cascade Attention-Guided Residue GAN framework (CAR-GAN) for the cross-modal audio-to-image translation task is proposed. 
	It explores cascaded attention guidance with a coarse-to-fine generation, aims at producing a more detailed synthesis from the jointly learned representation of both audio and image spaces.
	\item A novel residue module is presented, which is utilized to smooth the gaps between different modalities at class label space and is able to find correlations between different modalities. We also propose a new cascaded residue classification loss for more robust optimization. It not only helps the model generate more realistic images but also keeps the consistency between the two stages' generation processes.
	\item Qualitative and quantitative results demonstrate the effectiveness of the proposed CAR-GAN on the cross-modal audio-to-image translation task, and show state-of-the-art performance on the challenging Sub-URMP dataset~\cite{li2016creating} with remarkable improvements.
\end{itemize}

%% file: 2Related.tex
\section{Related work}
\label{related}

\par\noindent\textbf{Generative adversarial Networks (GANs)} is proposed by Goodfellow \etal{}~\cite{goodfellow2014generative}, complemented with adversarial method.
A vanilla GAN is composed of a generator and a discriminator. The discriminator is trying to discriminate whether an image is real or fake. Conversely, the generator is to learn to output images that can fake the discriminator. Since the GANs appeared, plenty of works such as~\cite{isola2017image,qiao2019mirrorgan,wang2018videotovideo,tang2020local} on computer vision are based on GANs. With the success of GANs, conditional GANs encode additional information as a reference into the GAN framework which will make sure the generator can run more straightforward to the target. cGANs have achieved remarkable results in image-to-image translation \cite{isola2017image,zhu2017unpaired,choi2018stargan,tang2020xinggan}, super-resolution image generation \cite{ledig2017photo,wang2018high} and style transfer \cite{huang2017arbitrary,chen2019gated}. 

\par\noindent\textbf{Image-to-Image Translation} adopts input-output data to learn a translation mapping between input and output domains. 
For instance, Isola \etal{}  propose Pix2Pix~\cite{isola2017image}, which is a general-purpose solution to image-to-image translation problems.
To further improve the quality of the generated images, works such as \cite{tang2019multi,zhang2018self,tang2020dual} try to employ the attention mechanism to force the generator to pay more attention to the distinguished content between different input and output domains.
In this paper, we embed the proposed attention mechanism into our cross-modal GAN model, which allows the generator to effectively pay attention to the most distinguished representations between audio and image modalities.
Moreover, previous works such as~\cite{shen2017learning,zhao2018learning} generate images using residual images which is different from ours, we employ the residual class-label to guide the generator for producing photo-realistic images. In this way, the generator only needs to focus on the high-level difference between the audio representation and image representation.

\par\noindent\textbf{Cross-Modal Learning}
represents any kind of learning that involves information obtained from more than one modality. Earlier work such as ~\cite{davenport1973cross,vroomen2000sound,kaiser2017one,zamir2018taskonomy} show cross-modal perception phenomena from the perspective of the neurological and psychological field. They try to figure out the mutual relation between auditory and visual information from a neurologist's or psychologist's view. Later, cross-modal multimedia retrieval starts booming since the revolution of multimedia technology. Works such as \cite{rasiwasia2010new,pereira2014role} take advantage of cross-modal learning to help retrieving. Afterwards, cross-modal learning is popular together with generative models, \eg{}, Variational Autoencoders (VAEs) \cite{spurr2018cross} and GANs \cite{park2018mc,hu2016now,am2019wav2pix}.

\par\noindent\textbf{Audio-to-Image Translation.} There are few works to address audio-to-image translation. Hao \etal{} propose a cycle encoder-decoder architecture to tackle the audio-to-image translation problem, under this framework, there are two sets of encoder-decoder, one for audio modality, and one for image modality. They try to solve the problem by making use of the cycle of the encoders and decoders. Besides, existing methods \cite{wan2018audio,chen2017deep,hao2018cmcgan} on audio-to-image translation take the advantage of latent Gaussian vector, where they design front convolution neural networks encoder to extract feature maps out of input audios. Later, the extracted feature maps are concatenated with the latent vector as new feature maps, the combined feature maps are fed into the generator to produce corresponding images, \ie{}, the latent vector plays an important role in this translation. 
However, different from these existing methods, we propose replacing the latent vector with the proposed residue class-label since it contains more meaningful representations between different modalities.

%% file: 3Approach.tex
\section{Cascade attention guided residue learning GAN}
\label{sec:approach}
In this section, we describe our proposed Cascade Attention-Guided Residue GAN (CAR-GAN) framework for cross-modal translation in detail. 
We start with the model formulation and then introduce the proposed objectives. Finally, we present the implementation details including network architecture and training procedure.
The overall framework of the proposed CAR-GAN is illustrated in Fig.~\ref{fig:framework}.

In stage one, we present a cascade attention guided generation sub-network, which utilizes both the audio signal $A$ and its class label $L_a$ as inputs to generate an image. 
The generated image $I'$ is further fed into the proposed residue module to obtain the corresponding image class label $L_i'$.
Next, we calculate the residual cross-modal label $L_r$ between the audio label $L_a$ and the image label $L_i'$. $L_r$ reflects the distance between the generated images and the real images in the semantic domain.

In stage two, the coarse synthesis $I'$ and the residual cross-modal label $L_r$ are combined as the input. In this way, the semantic difference between the generated images and audio signals, $L_r$ can be employed to guide the generator to further refine the generated image $I'$. As a result, the cross-modality semantic distance could be further reduced after the refinement.


\subsection{Stage I: Cascade attention guided generation}
\par\noindent \textbf{Class-Label Guided Generation with Self-Attention.}
Translating audio into an image is an extremely challenging task since it is difficult to tell any relationship between audio and image modalities directly.
To handle this challenge, previous works such as S2IC \cite{chen2017deep} and CMCGAN \cite{hao2018cmcgan} tried to employ a random Gaussian noise vector as input to guide the generator to produce a synthetic image.
We argue that the Gaussian noise vector will introduce some errors misguiding the generator.
Different from them, we input a more accurate audio class-label into the generator similar to \cite{choi2018stargan,tang2019dual}.
Specifically, as shown in~Fig.~\ref{fig:framework}, we first replicate $L_a$ spatially along with both height and weight dimension and then perform channel-wise concatenation with input LMS $A$ from audio space. Finally, we input them into the first generator $G_1$ and synthesize its corresponding coarse image $I'$ as $I'=G_1(A, L_a)$.
In this way, the audio class-label $L_a$ provides stronger supervision to guide cross-modal translation in the deep network.
Moreover, to force the generators to pay more attention to the most distinguished content between different modalities, we further introduce the self-attention mechanism into the generators. 
Zhang \etal{}~\cite{zhang2018self} proposed the Self-Attention Generative Adversarial Network (SAGAN) for image generation tasks.
Differently, in this paper, we propose a self-attention image-to-image translation network which allows long-range dependency modeling for cross-modal image translation task with drastic domain change.
Once the generators know which part they should pay attention to, the next goal is to generate images with more fine-grained details. Therefore, we cascade two generators and train them simultaneously.

\begin{figure*}[!t]
	\centering
	\includegraphics[width=0.8\linewidth]{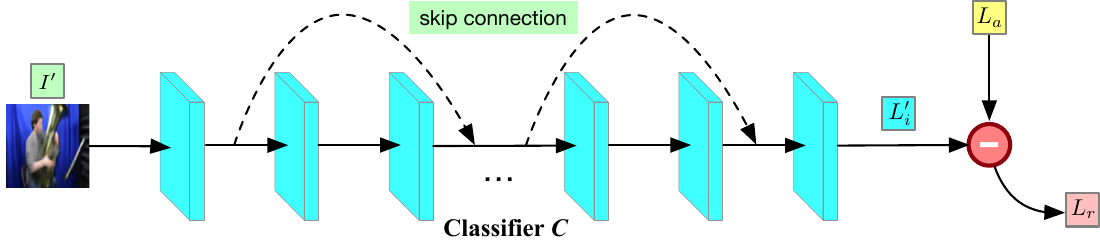}
	\caption{Residue Module: $I'$ denotes the output of the first generator $G_1$. The output of the classifier $C$ is $L_i'$, which is the predicted label of $I'$. 
	$L_a$ is the label of the audio, $L_r$ denotes the difference between two different modalities, \ie{}, audio label $L_a$ and image label $L_i'$. \textcircled{-} denotes channel-wise subtraction.}
	\label{fig:residue-model}
\end{figure*}

\par\noindent \textbf{Coarse-to-Fine Cascade Generation.}
Due to the complexity of the cross-modal audio-to-image translation task, we observe that the first stage generator $G_1$ only outputs a coarse synthesis with blurred artifacts, missing content, and high pixel-level dis-similarity. 
This thus inspires us to explore a cascade
generation strategy to boost the synthesis performance from the coarse predictions.
The Coarse-to-fine strategy has been used in different computer vision applications achieving promising performances, such as semantic segmentation~\cite{khan2017coarse} and object detection~\cite{pedersoli2015coarse,cristinacce2004multi}.
In this paper, we adapt the coarse-to-fine strategy to handle a more challenging audio-to-image translation task.
We observe significant improvement using the proposed cascade coarse-to-fine strategy, which is illustrated in the experimental section.

\subsection{Stage II:  Cross-modal residue label guided generation}

The overview of our proposed residue cross-modal label guided generation module is shown in Fig.~\ref{fig:residue-model}. This module consists of a pre-trained classifier to preserve the cross-modal label cycle consistency and a cross-modal residue label guided generation sub-network.

\par\noindent\textbf{Cross-Modal Label Cycle Consistency.}
Based on the theory \cite{choi2018stargan}, we expand it into our label consistency method. The coarse output $I'$ of stage I is fed into the classifier $C$ to generate an image classification label $L_i'$. To further reduce the space of possible mismatch between audio and image modalities, we hypothesize that the learned mapping functions should be cycle-consistent in cross-modal translation.
For the audio class label $L_a$, the translation
cycle should be able to bring it quite close to the image label $L_i'$,
\ie{}, $G_1(A, L_a) \rightarrow I'\rightarrow C(I') \rightarrow L_i' \approx L_a $.
We name this as cross-modal label cycle consistency since $L_i'$ is a label representation of the coarse output $I'$ in the image modality and $L_a$ is a label representation of the audio $A$ in the audio modality. 
Note that the proposed cross-modal cycle consistency is different from the cycle consistency in CycleGAN~\cite{zhu2017unpaired} which adapts the cycle-consistency between the input image and the reconstruction image in the image space, in this paper, we employ making two different modalities cycle consistent in the class-label space.

\par\noindent\textbf{Cross-Modal Residue Label Guided Generation.}
Previous works have shown that residual images can be effectively learned and used for image generation task.
For instance, Shen and Liu~\cite{shen2017learning} used the learned residual image as the difference between images before and after the face attribute manipulation.
Zhao \etal{}~\cite{zhao2018learning} trained networks to learn residual motion between the current and future frames for the image-to-video generation task, which avoids learning motion-irrelevant details.
Instead of manipulating the whole image, both approaches proposed to learn the residual images.
In this way, the manipulation can be operated efficiently with modest pixel modification.
However, in the paper, we propose the residue label rather than the residue image for the cross-modal image translation task.
Specifically, we first obtain the residue label $L_r$ between the image label $L_i'$ and the audio label $L_a$ by calculating $L_r=L_a-L_i'$.
Then we intend to generate the missing information $L_r$ in the second generator stage, which can be expressed as,
\begin{equation}
\small
\begin{aligned}
I''=G_2(I', L_r)=G_2(G_1(A, L_a), L_r).
\end{aligned}
\label{eq:residual}
\end{equation}
In this way, the generation process can be operated efficiently with modest pixel modification, \ie{}, the generator $G_2$ can flexibly preserve the similarities between the audio and image representations, and only model the differences between them.

\begin{figure*}[!t]
	\centering
	\includegraphics[width=0.7\linewidth]{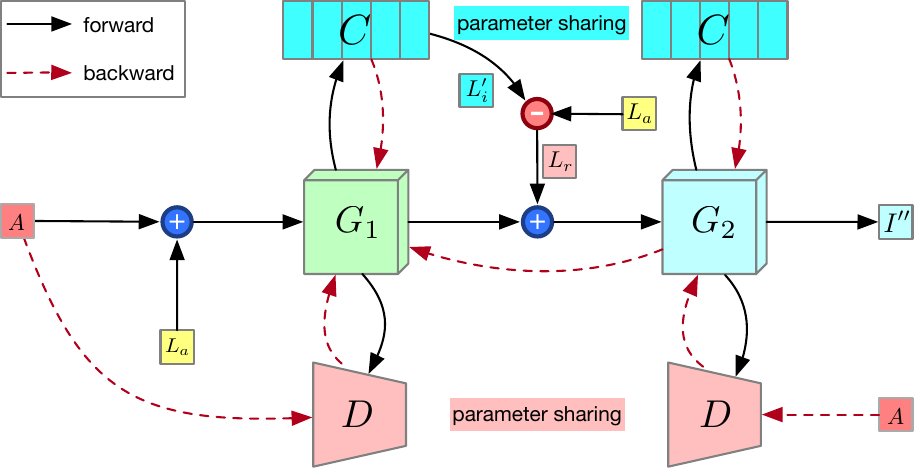}
	\caption{The backpropagation path of the proposed CAR-GAN. The solid line denotes the path for the forward process, the red dashed line denotes the backpropagation path. 
		We use the same pretrained classifier $C$ twice in our model, so does the discriminator $D$ but without pretrained.
		$I''$ denotes the final generated cross-modal image. \textcircled{+}, \textcircled{-} denote channel-wise concatenation and subtraction, respectively.}
	\label{fig:backward}
\end{figure*}

\subsection{Label consistency: Backpropagating via the classifier}
A vanilla cGAN conducts backpropagation mainly determined by the discriminator. The discriminator judges from the image-level information, but not the label-level semantic information.
We argue that different but corresponding modalities can match with the same semantic label.
Therefore, apart from performing backpropagation from the discriminator, we also backpropagate from the label classifier to make sure the generated images belong to the same label domain with input audios. 
During the training, the corresponding label, \ie{}, the instrument type of the input audio signals, is fed to the classifier. When we update the model during backpropagation, the parameters of the pre-trained classifier are fixed. Only the gradients of input images are passed back such that the images can be revised accordingly to match its semantic label. In this way, label consistency can be guaranteed.
The classifier is re-trained using a pre-trained model using ImageNet~\cite{imagenet_cvpr09}. 
The backpropagating path is described in Fig.~\ref{fig:backward}.
During the backpropagation of the classifier, we design a joint loss $\mathcal{L}_{C}$ for our two-stage generations. $\mathcal{L}_{C}$ is composed of two parts: the classification loss of stage I and stage II,
\begin{equation}
\small
\begin{aligned}
\mathcal{L}_{C} = \mathcal{L}(I',I'') = 
& \lambda_{I'}\mathcal{L}(I') + \lambda_{I''}\mathcal{L}(I''),
\end{aligned}
\label{eq:adv_1}
\end{equation}
where $\mathcal{L}$ is the cross-entropy loss function. $\lambda_{I'}$ and $\lambda_{I''}$ are coefficients to control the relative importance of the two objectives.

\subsection{Universal optimization objective}
\par\noindent\textbf{Adversarial Loss.} Our adversarial loss is composed of two parts since we adapt two stages generation. During the stage I, the adversarial loss of discriminator $D$ for differentiating generated audio-image pairs [$A$, $I'$] from real audio-image pairs [$A$, $I$] is formulated as following: 
\begin{equation}
\small
\begin{aligned}
\mathcal{L}_{cG_1}(A, I') = \mathbb{E}_{A, I} \left[\log D(A, I)\right] + \\
\mathbb{E}_{A, I'} \left[\log (1 - D(A, I'))\right],
\label{eq:adv_2}
\end{aligned}
\end{equation}
where $I$ is the ground truth image.
During the stage II, the adversarial loss of discriminator $D$ for differentiating generated audio-image pairs [$A$, $I''$] from real audio-image pairs [$A$, $I$] is formulated as following:
\begin{equation}
\small
\begin{aligned}
\mathcal{L}_{cG_2}(A, I'') = \mathbb{E}_{A, I} \left[\log D(A, I)\right] + \\
\mathbb{E}_{A, I''} \left[\log (1 - D(A, I''))\right].
\label{eq:adv_3}
\end{aligned}
\end{equation}
The above two adversarial losses both target to reduce the disagreements with ground truth images and generate more realistic synthesized images. Therefore, our adversarial loss is the total sum of the two stages:
\begin{equation}
\small
\begin{aligned}
\mathcal{L}_{cG} =  \lambda_{G_1}\mathcal{L}_{cG_1}(A, I') +  \lambda_{G_2} \mathcal{L}_{cG_2}(A, I'')
\end{aligned}
\label{eq:adv_4}
\end{equation}

\par\noindent\textbf{Universal Loss}. Besides the adversarial loss $\mathcal{L}_{cG}$, we also introduce the classification loss $\mathcal{L}_{C}$ and $L1$ loss function for better optimizing our CAR-GAN. Our final loss function is a combination of the three losses.
\begin{equation}
\small
\begin{aligned}
\min_{\{{G1}, {G2}, {C}\}} \max_{\{{D}\}} \mathcal{L} = \lambda_{c} \mathcal{L}_{C} + \lambda_{cG} \mathcal{L}_{cG} + \lambda_{L1}\mathcal{L}_{L1},
\end{aligned}
\label{eq:adv_5}
\end{equation}
where $\mathcal{L}_{L1}=\mathcal{L}_{L1}(I, I')+\mathcal{L}_{L1}(I, I'')$.
$\lambda_{c}$, $\lambda_{cG}$ and $\lambda_{L1}$ denote the trade-off parameters to control the significance of its corresponding loss function, respectively. Our model is trying to balance the min-max problem while training.

\subsection{Implementation details}
\par\noindent\textbf{Network Architecture.} Inspired by the work of Isola \etal{}~\cite{isola2017image}, we employ U-Net \cite{ronneberger2015u} as the backbone of our generators $G_1$ and $G_2$. U-Net is a Convolution Neural Network (CNN) architecture with skip connections between a down-sampling encoder and an up-sampling decoder, and it retains complex texture information of the input. 
We share the same network architecture in both generators. 
The convolutions of down-sampling layers and up-sampling layers are $4 {\times} 4$ kernel with stride 2 and padding 1. The filters in attention convolution layers are $1 {\times} 1$ with stride 1. 
For the discriminator $D$, we adapt PatchGAN as in \cite{isola2017image,zhu2017unpaired}. The kernel size of the attention convolution layers is also $1 {\times} 1$. Both the generators and discriminator have attention layers before the last two convolution layers. 
The other convolution layers in the discriminator have a kernel size of $4 {\times} 4$ kernel with stride 2 and padding 1. 
Batch normalization is used in our model. 
As for the classifier, we employ ResNet50 \cite{he2016deep} architecture which is pre-trained on the ImageNet. Then we add a fully connected layer at the end of the network and conducted transfer learning in the Sub-URMP dataset for high classification accuracy. The classifier is fixed while training our model.

\par\noindent\textbf{Training Details.} 
First, we employ preprocessing for every audio, where the audios are converted from waveform pattern to LMS pattern, which is a frequency warping pattern that allows for better representation of audio clip. After preprocessing all audios, we then input the LMS pattern into our proposed CAR-GAN. Moreover, the proposed CAR-GAN is trained and optimized in an end-to-end style.
$C$ is pre-trained and fixed while training.
We first train $G_1$, $G_2$ with $D$ fixed, and update parameters of $G_1$ and $G_2$ by the sum of gradients from $C$'s and $D$'s backpropagation, and then we train $D$ with $G_1$ and $G_2$ fixed, but the backpropagation of $C$ has no influence on the optimization of $D$, \ie{}, the optimization is only determined by $G_1$ and $G_2$. We apply Adam algorithm \cite{kingma2014adam} for optimizing both the generators ($G_1$, $G_2$) and discriminator ($D$) jointly. The betas of the Adam algorithm set to 0.9 and 0.999, respectively. Weights are initialized from a Gaussian distribution with standard deviation 0.2 and mean 0.

%% file: 4Experiments.tex
\begin{figure*}[!t]
    \centering
    \includegraphics[width=0.8\textwidth]{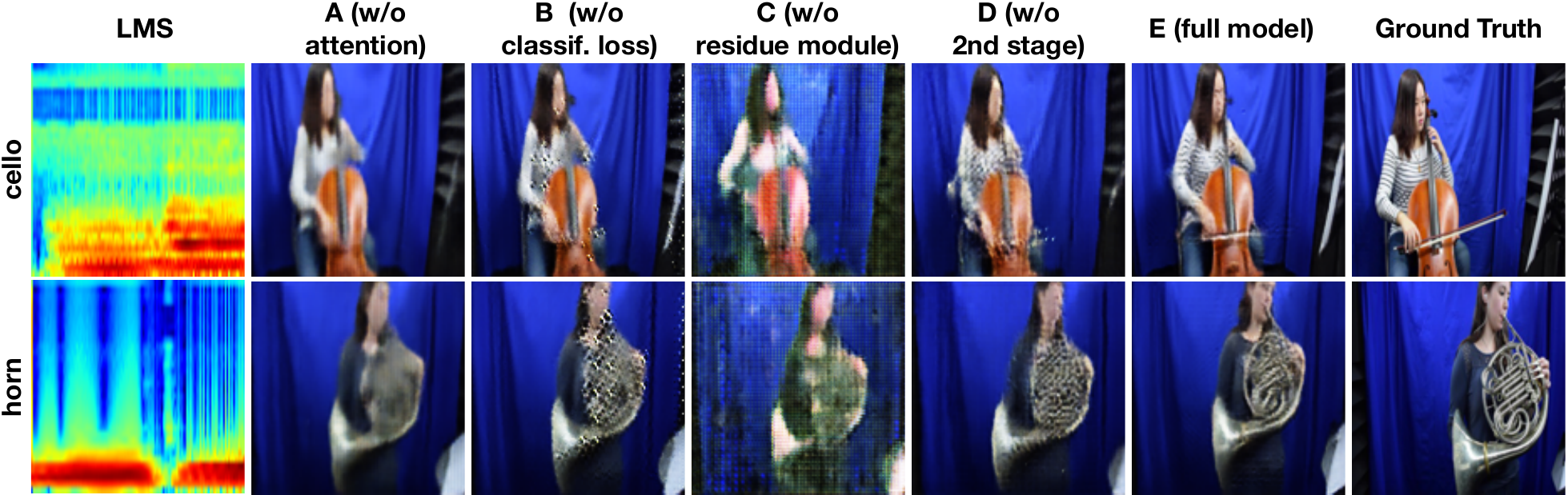}
    \caption{Ablation study: synthesized images by different models of the proposed CAR-GAN. LMS represents the LMS of the input audios.}
    \label{fig:ablation}
\end{figure*}

\section{Experiments}
\label{exp}
\subsection{Experimental settings}
\par\noindent\textbf{Datasets.} Following \cite{chen2017deep}, we adapt the widely used Sub-URMP (University of Rochester Musical Performance) dataset~\cite{li2016creating} to evaluate the proposed model. This dataset consists of 17,555 pairs of audios and images and has 13 kinds of instruments played by different people. It maps an image to half-second long audio, and the image is the first frame of the half-second long audio. 
\par\noindent\textbf{Parameter Settings.}
We resize images to $256 {\times} 256$ resolution as inputs. We implement with Pytorch and the experiments are running at 4 Nvidia GeForce GTX 1080 Ti GPUs with batch size 64. We set the learning rate to 0.0008, and stop our training at the epoch of 200. Both $\lambda_{I'}$ and $\lambda_{I''}$ in Eq.~\eqref{eq:adv_1} are set to 0.5. We set $\lambda_{G_1}$, $\lambda_{G_2}$ in Eq.~\eqref{eq:adv_4}, and $\lambda_{c}$, $\lambda_{cGAN}$ in Eq.~\eqref{eq:adv_5} all equal to 1, and $\lambda_{L1}$ equal to 100 in Eq.~\eqref{eq:adv_5}. 

\begin{table}[tbp]
    \centering
    \caption{Results of the proposed CAR-GAN for FID and IS metrics.}
    \scalebox{0.79}{
    \begin{tabular}{c|c|c|c|c|c|c} 
        \toprule
        Baseline   &    A & B & C & D & E & GT  \\
        \hline
        FID & 279.3022 & 332.1574 & 380.1104 & 307.3725 &  \textbf{207.3734} & -\\
        \hline
        IS & 3.2221 & 3.4337 & 2.0699 & 3.7215 &  \textbf{3.8180} & 4.7552\\
        \bottomrule
    \end{tabular}
    }
    \label{tab:ablation}
\end{table}

\begin{table}[tbp]
    \centering
    \caption{The classification accuracy of different methods.}
    \begin{tabular}{c|c|c}
        \toprule
        \multirow{2}{*}{Method} & \multicolumn{2}{c}{Accuracy} \\
        & Training       & Testing      \\ \hline
        S2IC                   & 0.8737         & 0.7556       \\ 
        \hline
        CMCGAN                 & 0.9105         & 0.7661       \\
        \hline
        Ours                   & \textbf{0.9954} & \textbf{0.9068}\\
        \bottomrule
    \end{tabular}
    \label{tab:accuracy}
\end{table}

\par\noindent\textbf{Evaluation Metrics.}
To compare with previous work \cite{chen2017deep,hao2018cmcgan}, we employ the classification accuracy as the metric, the only metric used in the previous two papers. The way we measure our model is that we first train a model with 99.56\% classification accuracy using ResNet50 trained on the Sub-URMP dataset. Specifically, the 99.56\% classification accuracy is obtained by training and testing on both the training and testing set. Then we test the pre-trained ResNet50 on our generated results. Following the same train/test split employed in Table 2 of \cite{chen2017deep}, our model is trained using the Sub-URMP dataset.
The intuition behind this is that if the generated images are realistic, the classifier trained on the real images will also achieve decent accuracy on the generated images during the testing stage. 
In addition to classification accuracy, we also employ Fr{\'e}chet Inception Distance (FID) \cite{heusel2017gans} and Inception Score (IS) \cite{salimans2016improved} metrics to further evaluate the fidelity of the generated images. Due to the lack of a pre-trained model and released code of previous work, we are not able to get FID and IS of their works.

\begin{figure}[tbp]
    \centering
    \includegraphics[width=0.8\linewidth]{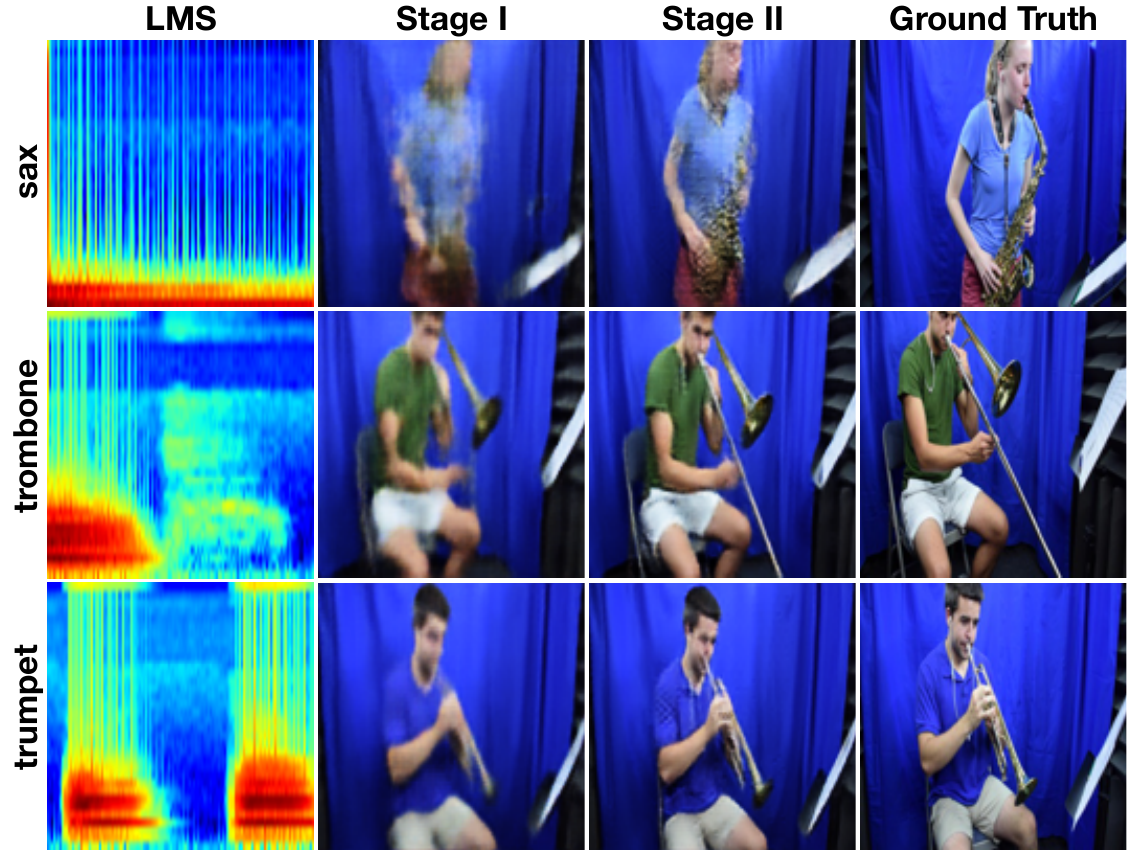}%
    \caption{Generated images of different stages of our model.}%
    \label{fig:stages}
\end{figure}

\subsection{Experimental results}
\par\noindent\textbf{Settings of Ablation Study.}
We perform ablation studies on the proposed CAR-GAN. 
We break down our model and assemble it into five different models. 
The following five models share a similar backbone, but a particular part is abandoned. Model A avoids using attention guided generations. Model B drops out the $\mathcal{L}_c$ loss. The residue module is taken out from model C and $L_r$ is replaced by $L_a$. Model D is running without stage II (no second generator). And the last model E is our fully proposed model. 
Fig.~\ref{fig:ablation} shows the corresponding cross-modal generated images of different models. Table~\ref{tab:ablation} depicts how models perform based on the metrics we employed. 

\par\noindent\textbf{Influence of Attention Mechanism.}
Compared with our proposed full model E, model A which avoids the attention mechanism performs slightly worse. The results of model A shows the overall contour of images, but some details are left out. That is, the attention mechanism does enhance the representation ability of our model.

\begin{figure}[tbp]
    \centering
    \includegraphics[width=0.87\linewidth]{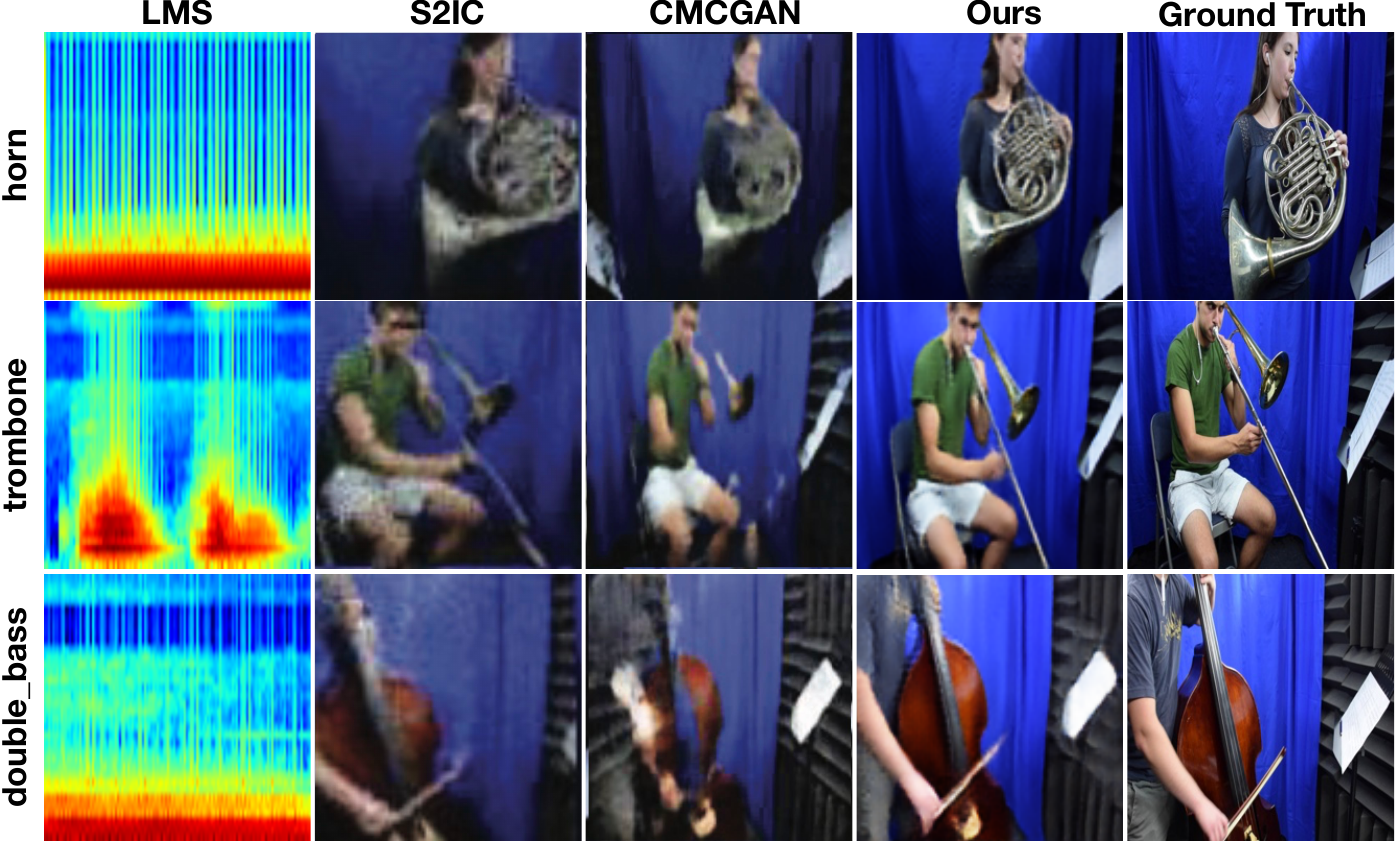}
    \caption{Synthesized images of different methods on the Sub-URMP dataset.}%
    \label{fig:comparisons}
\end{figure}

\begin{figure}[tbp]
    \centering
    \vspace{-0.2cm}
    \includegraphics[width=0.8\linewidth]{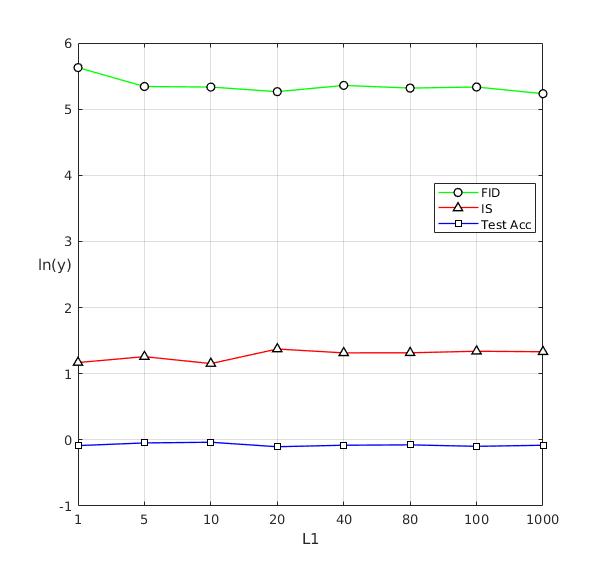}
    \caption{Impact of L1 regularizer. $y$ denotes the performance for each metric. The figure shows how the metrics' value changing with different weight of L1 regularizer in loss function. We adapt $ln$ function on $y$ for  better visualization.}
    \label{fig:L1}
\end{figure}
\par\noindent\textbf{Influence of Classification Loss.}
If we add our proposed classification loss into the model, we make an improvement by 37.56\% in FID and 11.19\% in IS since the classification loss $\mathcal{L}_c$ awards the generators strong guidance towards the ground-truth. This tells generators from an overall classification view. Therefore, the generators know the appropriate direction to go from the label domain.

\par\noindent\textbf{Influence of Residue Module.}
With our residue module, we achieve an amazing improvement on FID by 45.44\% and on IS by 84.45\%. Our residue module supplements the missing information during the generation of Stage I, making that the generated images belong to the same domain and keep balance in the label domain between inputs and outputs. 

\par\noindent\textbf{Influence of Two-Stage Generation.}
Our two stages of generators lead to the improvement of 32.53\% in FID and 2.59\% in IS. The single generator has weak representation ability for a complicated cross-modal generation. Thus, the union of two or more generators can progressively improve representation ability and performance. To visualize the influence of the two-stage generation, we display synthesized images by different stages in Fig.~\ref{fig:stages}. The generated images by the second stage have more fine-grained details and are more realistic, \ie{}, it certifies that the two-stage generation is beneficial for the whole generation procedure.

\begin{figure}[tbp]
    \centering
    \includegraphics[width=0.8\linewidth]{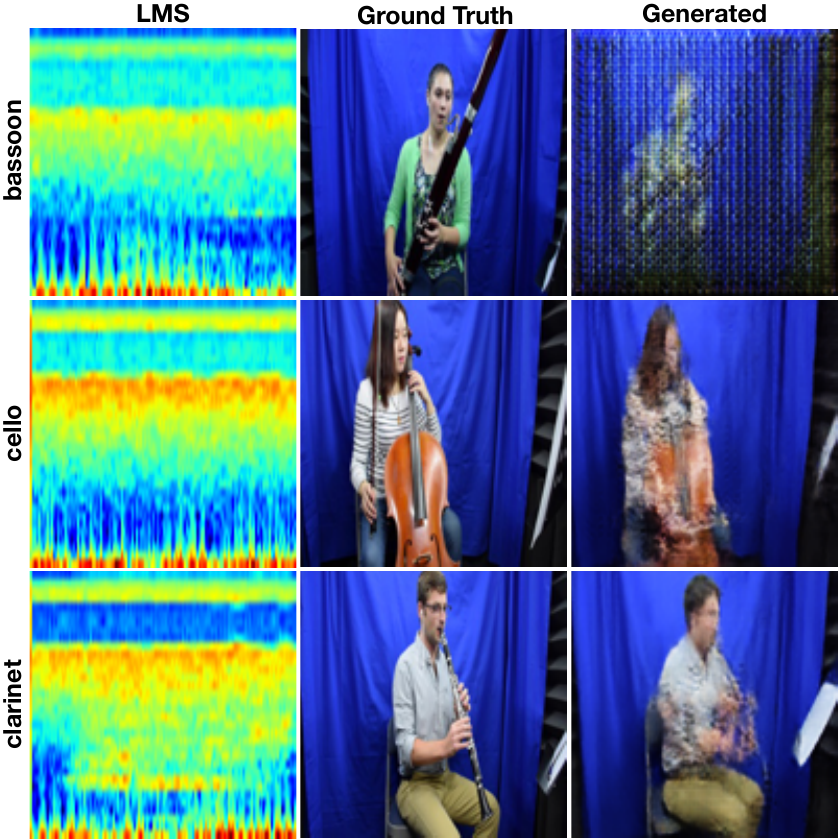}%
    \caption{Failure cases during inference on the Sub-URMP dataset.}%
    \label{fig:failed}
\end{figure}

\par\noindent\textbf{Impact of L1 regularizer.}
To show the model is not overfitting by L1 loss, a more detailed experiment is performed to evaluate the impact of the structural L1 regularizer for the generated images. To better show the different results, we plot the trajectory of the metrics’ value in Fig.~\ref{fig:L1}.
\par\noindent\textbf{State-of-the-art Comparisons.} We show the quantitative and qualitative results with comparison methods in Table~\ref{tab:accuracy} and Fig.~\ref{fig:comparisons}. We make an improvement in training accuracy by 13.93\%, 9.32\% compared to S2IC~\cite{chen2017deep} and CMCGAN~\cite{hao2018cmcgan}, respectively. As for the testing, we achieve an increase of 20.01\% and 18.37\% compared with S2IC and CMCGAN. Furthermore, we produce more realistic and detailed images compared with the previous methods as illustrated in Fig.~\ref{fig:comparisons}.

\par\noindent\textbf{Failure Cases and Analysis.}
During our experiments, we find there are some synthesized images like Fig.~\ref{fig:failed} which are randomly combined by learned features. Moreover, when we carefully look into these cases, we finally find out that the corresponding inputs of these failure cases are more like noises that are randomly distributed, and the GT images show people mostly hold the instrument still and wait, that is, there is no sound making by instruments, the audios mainly consist of background and other noises.

%% file: 5Conclusion.tex
\section{Conclusion}
\label{con}
In this paper, we design a novel Cascade Attention Guided Residue Learning GAN (CAR-GAN) to solve the challenging cross-modal audio-to-image translation task. Particularly, it employs cascaded attention guidance and a coarse-to-fine generation strategy. A novel residue learning model is also proposed to tackle the cross-modal class-label dis-match problem between audio and image modality. By introducing the residue module, generators learn to produce residue features between two stages, which pushes the output closer to its corresponding real image in a high-level semantic space. Finally, the proposed joint classification loss facilitates the model generation and keeps consistency in the label domain. Experimental results show the state-of-the-art performance on the cross-modal translation task.

\noindent \textbf{Acknowledgement:} This research was partially supported by NSF NeTS-1909185 and a gift donation from Cisco Inc. This article solely reflects the opinions and conclusions of its authors and not the funding agents.